\documentclass[letterpaper, 10pt, conference]{ieeeconf}      

\IEEEoverridecommandlockouts                              
\usepackage{graphicx}
\usepackage{subcaption}
\usepackage{amsmath}
\usepackage{algorithm}
\usepackage{booktabs}
\usepackage[dvipsnames,svgnames,x11names]{xcolor}

\usepackage[noend]{algpseudocode}
\usepackage{listings}
\def\BibTeX{{\rm B\kern-.05em{\sc i\kern-.025em b}\kern-.08em
    T\kern-.1667em\lower.7ex\hbox{E}\kern-.125emX}}
\usepackage{stfloats}
\usepackage{float}
\usepackage{cite}
\usepackage[hidelinks]{hyperref}

\usepackage[dvipsnames,svgnames,x11names]{xcolor}
\usepackage[final]{changes}

\begin{document}




\title{\LARGE \bf
Anticipatory Planning: Improving Long-Lived Planning by Estimating Expected Cost of Future Tasks
}

\author{Roshan Dhakal, Md Ridwan Hossain Talukder and Gregory J. Stein%
\thanks{Roshan Dhakal, Md Ridwan Hossain Talukder and Gregory J. Stein are affiliated with the CS Department, George Mason University, Fairfax, VA: {\tt\small \{rdhakal2, mtalukd, gjstein\}@gmu.edu}}}%

\maketitle
\thispagestyle{empty}
\pagestyle{empty}
\begin{abstract}
We consider a service robot in a household environment given a sequence of high-level tasks one at a time. Most existing task planners, lacking knowledge of what they may be asked to do next, solve each task in isolation and so may unwittingly introduce side effects that make subsequent tasks more costly. In order to reduce the overall cost of completing all tasks, we consider that the robot must anticipate the impact its actions could have on future tasks. Thus, we propose \emph{anticipatory planning}: an approach in which estimates of the expected future cost, from a graph neural network, augment model-based task planning.  Our approach guides the robot towards behaviors that encourage preparation and organization, reducing overall costs in long-lived planning scenarios. We evaluate our method on blockworld environments and show that our approach reduces the overall planning costs by 5\% as compared to planning without anticipatory planning. Additionally, if given an opportunity to \emph{prepare} the environment in advance (a special case of anticipatory planning), our planner improves overall cost by 11\%.

\end{abstract}

\section{Introduction}



\begin{figure}
    \vspace{5pt}
    \includegraphics[width=8.5cm]{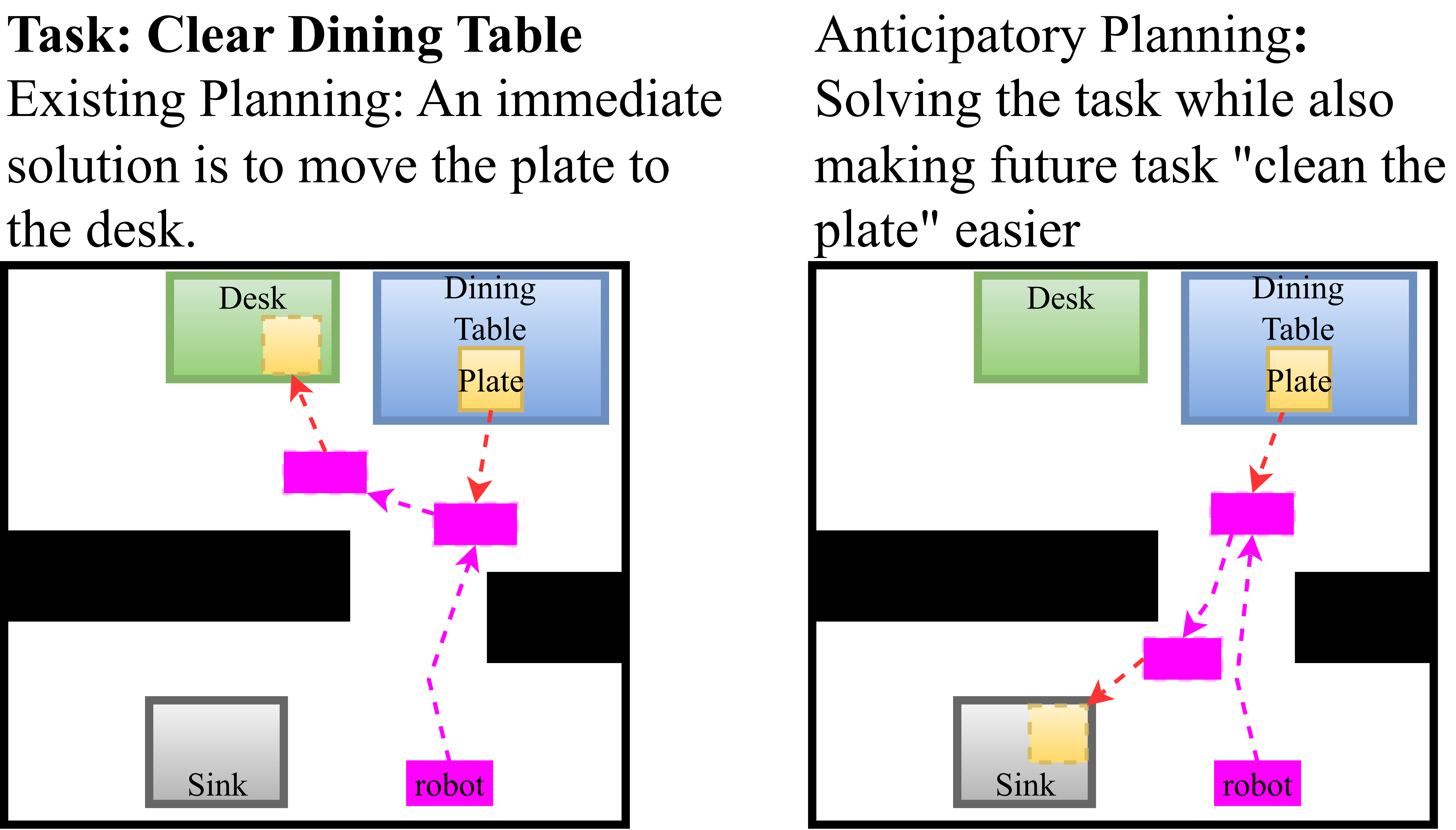}
    \centering
    \caption{\textbf{An Anticipatory Planning Scenario}: The immediate plan to clear the dining table (left) is to move away the plate to the desk with a minimum cost. However, this limits the use of the desk later and also increases the cost of cleaning the bowl later. Our robot instead anticipates how it may later interact with the environment and takes preemptive action (right) to reduce expected future cost.}
    \centering
    \label{fig:example1}
    \vspace{-15pt}
\end{figure}

We consider a service robot in a household that is given a sequence of high-level task specifications (task planning problem) one at a time, each expressed in Planning Domain Definition Language (PDDL)~\cite{fox2003pddl2} by a human operator. Most existing task planners~\cite{garrett2020pddlstream, plaku2010sampling,toussaint2015logic,kaelbling2011hierarchical, srivastava2014combined,kim2019learning,chitnis2016guided,dantam2016incremental} in this setting solve each task objective one at a time in isolation to complete each in minimum cost. However, since the environment persists between tasks, the robot may unwittingly introduce problematic side effects for the subsequent tasks it has yet to be assigned. For example (Fig.~\ref{fig:example1}), consider a household robot given a task from a sequence of tasks to clear the dining table that has a plate on it. Though the quickest solution is to move the plate to the desk, this limits the use of the desk later and increases the planning cost of a future task to clean the bowl, and so should instead be placed in the kitchen sink. 

If the robot is to minimize an overall cost over \emph{the entire sequence of tasks}, it must have the capacity to understand the impact of its actions---those that solve the current task---on future tasks. In other words, to reduce the overall cost, it would require the robot to act in a way that seeks to pay down both the immediate cost of completing a current task and the expected cost associated with future tasks in the sequence which we call the \underline{\textit{anticipatory cost}}. We introduce \underline{\textit{anticipatory planning}}: planning so as to minimize the joint objective of planning cost for the current task and the anticipatory cost.


If the robot were provided all its tasks in advance, existing task planners could be used to find the minimum-total-cost plan.
Though the robot will not typically know future tasks in advance, we assert that there exists a \emph{task distribution} inherent to the environment that can be used to define the \emph{anticipatory cost}: the \emph{expected} future cost.
To plan well, the robot must determine which actions will reduce the joint cost of completing the current task and the anticipatory cost. Given access to task distribution and sufficient computation, the robot could reduce anticipatory cost. However, during deployment, computing anticipatory cost is often computationally infeasible, as there could be hundreds of possible future tasks, or because the robot lacks direct access to the task distribution when deployed in its new environment.




So as to overcome the challenges associated with limited knowledge or computation, we use learning to estimate the anticipatory cost.
Doing so allows us to transfer experience from an offline training phase---when we have both sufficient computation and direct access to the task distribution---to improve planning performance during deployment.
Offline, the robot has direct access to the task distribution, from which it generates data to train a Graph Neural Network~\cite{scarselli2008graph,battaglia2018relational, kipf2016semi} to estimate the anticipatory cost. During deployment, the robot uses this \emph{anticipatory cost estimator} to augment model-based planning and improve overall cost.

\begin{figure}
    \vspace{5pt}
    \includegraphics[width=8.5cm]{figures/prep_example.pdf}
    \centering
    \caption{\textbf{Preparation as Task-Free Anticipatory Planning}: The left figure shows the current state of an environment with probable future tasks. In the right figure, our robot anticipates future tasks and prepares an environment to minimize the overall planning cost of a future task.}
    \centering
    \label{fig:example2}
    \vspace{-15pt}
\end{figure}

\deleted[id=RD]{Anticipatory Planning enables robots to take action in an anticipation of future tasks so that when a task is given, the planning cost is reduced.} Even when a task is not given, our Anticipatory Planning approach allows our robot to ready the environment in anticipation of future tasks, reducing cost for when a task is eventually provided. We refer to this task-free planning setting as \replaced[id=RD]{\emph{preparation}}{preparation}, illustrated in Fig.~\ref{fig:example2}.

\replaced[id=RD]{To evaluate Anticipatory Planning,}{We evaluate our Anticipatory Planning approach} averaged across an ensemble of multi-task sequences in a blockworld environment. We demonstrate that our approach improves overall cost by 5\%. We also demonstrate that even in the absence of an immediate task, anticipatory planning allows the robot to \textit{prepare} the environment and reduce the overall costs of potential future tasks by 11\%.

\section{Related Work}
\textbf{Task Planning and Learning-Augmented Planning:} Task and Motion Planning (TAMP) involves jointly solving a high-level task and the motion planning to accomplish those tasks~\cite{garrett2021integrated}.\footnote{We do not currently tackle \emph{Integrated} Task and Motion Planning. As we highlight in Sec.~\ref{pddl}, low-level motion planning and the costs of motion-borne actions are computed independently of high-level task planning.}
Many existing works in this space~\cite{ garrett2020pddlstream, plaku2010sampling, toussaint2015logic, kaelbling2011hierarchical, srivastava2014combined, dantam2016incremental} are focused on solving a TAMP problem as quickly as possible by using so-called classical planning strategies.

Recently, learning has been used to accelerate planning, extending the limits of plan complexity and improving plan quality.
Learning techniques have been integrated into many aspects of TAMP systems, from learning samplers for continuous values~\cite{kim2019learning, chitnis2016guided, wang2018active,chitnis2019learning} to learning guidance for symbolic planning~\cite{kim2020learning}. However, approaches for both \textit{classical TAMP} and \textit{learning for TAMP} primarily focus on solving small-scale or deterministic planning problems and are not directly applicable to solving our anticipatory planning objective.

\textbf{Learning from Demonstration and Common Sense Planning:} 
Common sense reasoning describes a class of scenarios in which robot behavior is expected to agree with human intuition, though so far has generally defied precise mathematical definition. Agrawal~\cite{agrawal2021taskspec} identifies under-specification in the task definition as limiting common sense and points to approaches to transfer and meta-learning~\cite{finn2017maml,chen2020selfsuper} or diverse skill learning~\cite{eysenbach2018diversity} that have shown limited promise in addressing this under-specification.
In the context of service and assistive robots, approaches such as learning from demonstration (LfD), inverse reinforcement learning~\cite{kim2021imitateplan}, reward learning~\cite{jeon2020reward}, and similar~\cite{shah2021benefits} have proven effective in situations where an explicit reward function can be difficult to write down, encouraging intrinsic motivation~\cite{kulkarni2016intrinsic,mahzoon2019ostensive}, courtesy~\cite{choudhury2019utility,sun2018courteous}, and avoiding side effects~\cite{shah2019preferences,krakovna2020avoiding} in service-like domains.
These approaches to improve robot behavior typically struggle for long-horizon planning in non-deterministic settings~\cite{ravichandar2020lfdsurvey} and so efforts in this space are primarily limited to deterministic settings---for low-level~\cite{whitney2018learning,moro2018learning,pastor2011skill,vandenBerg2010lfdsurgery} and high-level~\cite{manschitz2014learning,konidaris2012robot} planning and from physical corrections~\cite{li2021learning,meszaros2022learning}---or short-time-horizons~\cite{choi2018uncertainty,thakur2019uncertainty}.


\textbf{Graph Neural Networks and Applications to TAMP:} In this work, we use graph neural networks (GNNs)~\cite{scarselli2008graph, kipf2016semi, hamilton2017inductive}. See~\cite{battaglia2018relational} for a recent survey. Recently, GNNs have proven effective for geometric integrated TAMP problems. Graph representations of the state of an environment can be fed into a graph neural net to guide sample-based motion planning~\cite{kim2020learning} or to estimate the relative importance of objects in a scene to accelerate search over tasks~\cite{silver2021planning}. 
We rely on the powerful representational capacity and generalization capabilities of GNNs to estimate the anticipatory cost in an effort to improve planning across multiple tasks.

\section{Problem Statement}

\subsection{Anticipatory Planning}
We want to reduce the overall cost of solving a sequence of $N$ tasks $\tau$, given only one task at a time considering both the immediate cost of completing the current task and the anticipatory cost of a state for future tasks in the sequence. If we were to know the sequence of tasks in advance, we could find a plan that minimizes total cost according to
\begin{equation} \label{eq:full-problem}
\begin{split}
s^*_{g_1}, \cdots\!,  s^*_{g_N} & = \!\!\!\!\!\! \underset{{s_{g_i}\in G_{\tau_i}} \forall i \in {1 \cdots n}}{\text{argmin}}\!\!\!\!\!\! \left[\substack{V^*_{s_{g_1}}(s_0)+V^*_{s_{g_2}}(s_{g_1}) + \cdots + V^*_{s_{g_n}}(s_{g_{n-1}})}\right]\\
\end{split}
\end{equation}
where $\tau_i$ is $i^{th}$ task in sequence, $V^*_{s_{g_{i+1}}}(s_{g_{i}})$ is the minimum cost of completing $\tau_i$, which can be computed via any deterministic planner, for example, Sec.~\ref{pddl}. State $s_{g_i}$ is the final state of completing $\tau_i$ and also the initial state for completing task $\tau_{i+1}$. 
However, since the robot will in general not know all the tasks in advance, it instead seeks a policy that minimizes \emph{expected} total cost, where future tasks $\tau$ are drawn from a \emph{task distribution}: $\tau \sim P(\tau)$.

Reasoning about future cost over a long sequence of tasks is computationally challenging, and so we instead seek to minimize the expected cost associated with an immediately available task and a \emph{single} next task in the sequence.
In general, planning in this domain involves first moving from starting state $s_0$ to some intermediate goal state $s'$, belonging to the set of goal states $G_{\tau}$ in which the task $\tau$ is completed, and then subsequently completing another task $\tau' \sim P(\tau)$. 
We refer to this planning objective as \underline{\emph{anticipatory planning}}, the aim of which is to find the intermediate goal state $s^*_g$ that minimizes the total cost according to

\begin{equation}\label{eq:antplan}
\begin{split}
s^*_g & = \underset{s'_g\in G_{\tau}}{\text{argmin}} \Bigg[V^*_{s'_g}(s_0) + \underset{\tau'} \sum P(\tau') V^*_{\tau'}(s'_g)\Bigg]\\
 & = \underset{s'_g\in G_{\tau}}{\text{argmin}} \Bigg[V^*_{s'_g}(s_0) + V^*_{A.P.}(s'_g)\Bigg]
\end{split}
\end{equation}
where $V^*_{s'_g}(s_0)$ is the optimal cost of moving from state $s_0$ to state $s'_g$, and $V^*_{\tau'} (s'_g)$ is the cost of completing task $\tau'$ starting from state $s'_g$. We refer to $V^*_{A.P.}(s'_g)$ as the \emph{anticipatory planning cost}: the expected cost of completing a single follow-up task starting from state $s'_g$.
\begin{equation} \label{eq:apcost}
V^*_{A.P.} (s) = \underset{\tau'} \sum P(\tau') V^*_{\tau'}(s)
\end{equation}
While $V^*_{s'_g}(s_0)$ can be computed using existing task planning solvers, the anticipatory planning cost is often too difficult to compute online, and so we instead estimate it via learning (see Sec.~\ref{sec:learning}).


\subsection{Preparation as Task-Free Anticipatory Planning}\label{sec:theory:prep}
Even when the robot is not assigned an immediate task $\tau$---a common scenario for household robots, which may not be constantly given active instruction---anticipatory planning can inform how the robot can reconfigure its environment so as to reduce expected future cost for when it is eventually given a task.
We refer to this objective as $\emph{preparation}$, in which the robot seeks to find a prepared state of the environment $s^*_{prep}$, when the state has the minimum expected cost $V^*_{A.P.}$.
This scenario is related to the anticipatory planning objective of Eq.~\eqref{eq:antplan}, yet the robot (not given an immediate task) accumulates no cost for immediate actions.
As such, it is the robot's objective to minimize only the anticipatory planning cost $V^*_{A.P.}$:
\begin{equation}\label{eq:prep}
s^*_{prep} = \underset{s \in S}{\text{argmin}} \left[ V^*_{A.P.} (s) \right]
\end{equation}
where $S$ is the set of all possible states in the environment.

In our experiments, we will demonstrate the effectiveness of both anticipatory planning and preparation in improving the expected cost of planning.

\begin{figure}
    \vspace{7pt}
    \includegraphics[width=8.5cm]{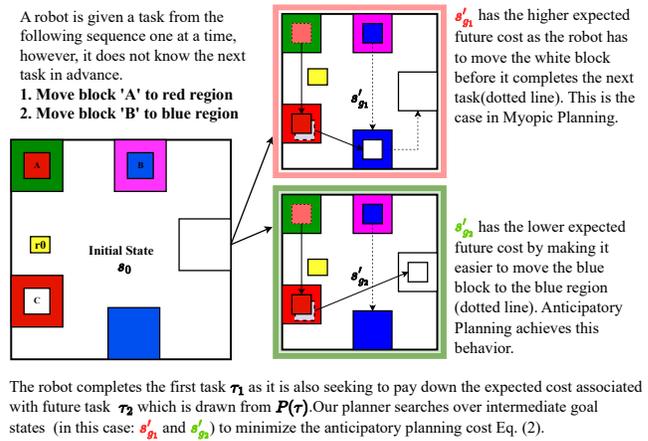}
    \centering
    \caption{A schematic overview: Our robot reduces the overall planning cost of the sequence by considering the immediate cost of the current task and the expected cost associated with future tasks.}
    \centering
    \label{fig:schematic}
    \vspace{-20pt}
\end{figure}

\section{Preliminaries: Task Planning with PDDL} \label{pddl}

In this paper, we consider pick-and-place-style task planning problems, for which action costs are computed separately via low-level motion planning.
To achieve high-level goals---e.g., clearing a table---robots need to be able to carry out high-level actions, such as picking up a bowl and moving around the environment. We represent both the tasks and the actions available to the robot for this Task Planning problem using the Planning Domain Definition Language (PDDL)~\cite{fox2003pddl2}. PDDL defines a fully observable deterministic planning problem as tuple $(A, s_0, g)$, where $A$ is a set of parameterized actions, $s_0$ is an initial state and $g$ is a goal condition for the problem. An example PDDL definition of a \texttt{pick} action is included in Listing~\ref{lst:pddl}.

\begin{lstlisting} [ basicstyle=\small, linewidth=\columnwidth,breaklines=true,frame=tb, caption=PDDL code definition for action \texttt{pick}, label={lst:pddl}, float]
  (:action pick
    :parameters (?r ?b ?s)
    :precondition (and (Robot ?r) (HandEmpty ?r) (At ?r ?s) (In ?b ?s))
    :effect (and (Holding ?r ?b) (not (In ?b ?s))
                 (not (HandEmpty ?r))
                 (increase (total-cost) 100) 
                 ))
\end{lstlisting}
Actions involving picking up or placing an object have a constant cost.
By contrast, costs for the $move$ action come from a separate motion planning process. \deleted[id=RD]{We are only doing high-level task planning, however, in order to get the cost of planning for a given task, we also take the robot’s motion into account, therefore, we use a motion planner to get the cost.}
We consider a $move$ action as the robot's motion from one location $l1$ to another $l2$, represented by the tuple $(l1, l2, f)$ where $f$ is a collision checking function between $l1$ and $l2$.
We use the Lazy Probabilistic Roadmap (Lazy PRM) algorithm~\cite{bohlin2000path} to get the cost for $move$ action provided by~\cite{garrett2020pddlstream}. Lazy PRM operates like traditional PRM~\cite{kavraki1996probabilistic}, but lazily checks the collision, which removes redundant sampling of regions that does not provide the solution and accelerates planning.

We use Fast Downward~\cite{helmert2006fast} for task planning, using A* search with the admissible \texttt{max} heuristic, so as to yield optimal plans.\footnote{We note that one could use an inadmissible search heuristic---e.g., the well-known Fast-Forward (FF) heuristic~\cite{Hoffmann2001FFTF}---for both data generation and planning and trade optimality for improvements in planning speed. In this work, we focus only on admissible search via A*.} The planner consumes in the PDDL definition of the problem, including pre-computed motion planning costs for \texttt{move} actions, and returns minimum-cost plans that solve the task. We use the notation $V^*_{s_g} (s_0)$ to mean ``the cost of the optimal plan to get from state $s_0$ to goal state $s_g$.''

\section{Estimating Anticipatory Cost}\label{sec:learning}

Direct computation of the anticipatory cost $V^*_{A.P.}(s)$ during deployment is often not feasible---either because it is computationally challenging or because the robot does not have direct access to the task distribution $P(\tau)$. This makes the robot unable to search over the states during planning to solve Eq.~\eqref{eq:antplan}, which depends on the anticipatory cost $V^*_{A.P.}(s)$.
To overcome this limitation, we instead use learning to estimate the anticipatory cost of the state $s$. Offline, we randomly generate task sequences drawn from the task distribution $P(\tau)$, with which we compute samples of the anticipatory cost to serve as training data.

We learn the anticipatory cost in a supervised manner. 
To be useful as a component in a search procedure, we require a process for mapping states to inputs to our estimator. During deployment, a photorealistic simulator may not be available, so images are not well-suited for this purpose.
Instead, we represent the environment state as a graph and train a \emph{Graph Neural Network}~\cite{battaglia2018relational} to estimate the anticipatory cost.
Graph Neural Networks (GNNs) have proven effective tools for making predictions in household robotics settings~\cite{kim2020learning,silver2021planning}.

\subsection{Representing the Environment State as a Graph}
\label{sec:learning:representation}
The state $s$ of an environment $\mathcal{E}$ defines the location of each object in that environment as well as the location of various regions in which those objects can be placed.
Each graph $\mathcal{G}$ represents a state. 
For example, in the blockworld environment, as shown in Fig.~\ref{fig:schematic}, the state is defined as blocks being placed in specific regions.

\begin{figure}
    \vspace{6pt}
    \hspace{1pt}
    \includegraphics[width=8.5cm]{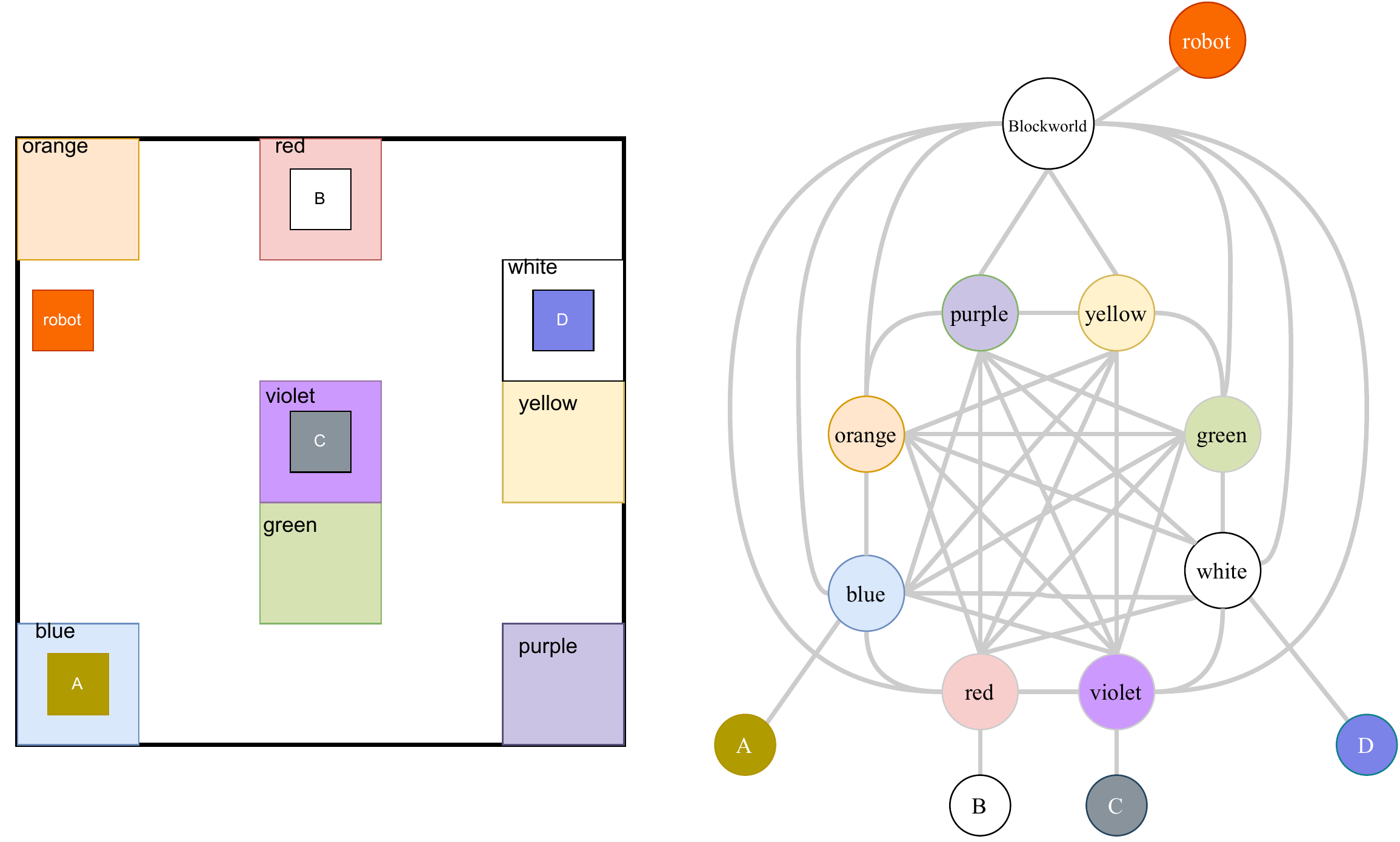}
    \caption{An example of a state in an environment and its corresponding graph structure for training data}
    \centering
    \label{fig:stategraph}
    \vspace{-10pt}
\end{figure}

To represent the environment state as a graph $G$, nodes correspond to (i) places in the environment (including a node for the environment itself), (ii) objects in the environment, and (iii) locations in which objects can be placed. Edges are created whenever one entity is contained by another. For example, a block is in the red region, so an edge connects the two. An illustration of an example state and its corresponding graph can be seen in Fig.~\ref{fig:stategraph}.

To act as input to a graph neural network, nodes have accompanying \emph{node input features} that include properties of the entity it represents. Node features in our graph consists of a one hot encoding of its entity type (whether it is a room, location, or object), entity color (specifying its semantic class), and the 2D coordinate of its location in the environment. For example, a node for a \texttt{book} from Fig.~\ref{fig:stategraph} has the entity type is \texttt{object}, thus we get one hot encoding of \texttt{object}. Since it is not both room and location type, all vectors in both encodings are zeros. \texttt{Book} is an object type from which we get the corresponding encoding. It also has a color \texttt{blue} which has scalar value $(0.0, 0.0, 1.0, 1.0)$.


\subsection{Data Generation}
\label{sec:learning:data}

To train our graph neural network (GNN) to estimate the anticipatory cost, we require training data generated from each environment class of interest. To obtain this data, we use Sec.~\ref{pddl} to solve each planning problem instance from task distribution $P(\tau)$ from randomly sampled 200 states in each seed of environment. We generate data from 250 procedurally-generated training environments and 150 distinct testing environments. We compute the anticipatory cost $V_{A.P.}$ for each state $s$. Each labeled datum consists of the graph $\mathcal{G}$ of $s$ and its anticipatory cost $V_{A.P.}$. 

\subsection{Graph Neural Network Structure and Training}
\label{sec:learning:gnn}
Our estimator is a graph neural network that takes as input a graph $\mathcal{G}$ with nodes $v$ and edges $e(v1, v2)$ corresponding to the environment state $s$ trained to estimate the anticipatory cost $V^*_{A.P.}(s)$.
Our graph neural network consists of three SAGEConv layers~\cite{hamilton2017inductive}, provided by the PyTorch Geometric package~\cite{pyg}, each followed by leaky ReLU activation function. We also add a mean pooling layer for batch-wise graph level outputs averaging node features and, finally, a linear classifier layer to produce the anticipatory cost. We use mean absolute error (MAE) as our loss function. We train our model using Adam optimizer~\cite{kingma2014adam} on PyTorch~\cite{pytorch}
 with a batch size of 8 and roughly using 50k training samples. We train for 10 epochs using a fixed learning rate of 0.01.







\section{Search In Anticipatory Planning}\label{sec:search}
\begin{figure}[t]
\vspace{1em}
\begin{algorithm}[H]
\caption{Search for an Intermediate State}\label{alg:cap}
\begin{algorithmic}[1]
\Function {search}{$s_0, \textsc{Estimator}, \tau$}
\State $\pi, V^*_{\tau}(s'_g) = \textsc{TaskPlan}(s_0, \tau)$
\State $s^* = s'_g$
\State $V^*_\text{total} (s^*) = V^*_{s'_g}(s_0) + \textsc{Estimator}(s'_g)$
\State $G_{\tau} = \textsc{SetOfAlternateGoalStates}(\tau, \pi)$
\For {$s' \in G_{\tau}$}
\State $V^*_\text{total} (s') = V^*_{s'}(s_0) + \textsc{Estimator}(s')$
\If {$V^*_\text{total} (s') \leq V^*_\text{total} (s^*) $}
\State $V^*_\text{total} (s^{*}) = V^*_\text{total} (s') $
\State $s^* = s'$
\EndIf
\EndFor
\State \Return $s^{*}$
\EndFunction
\end{algorithmic}
\end{algorithm}
\end{figure}
It is computationally infeasible to enumerate all states in a complex environment when searching during anticipatory planning to find states that reduce overall cost according to Eq.~\eqref{eq:antplan}. 
Instead, we rely on an iterative procedure to search more efficiently over states likely to improve the total cost, shown in Alg.~\ref{alg:cap}. 

We first compute the minimum-cost plan for task $\tau$ using the solver, $\textsc{TaskPlan}$ (Sec.~\ref{pddl}), which returns both the plan $\pi$ and the immediate cost of that plan $V^*_{\tau}(s'_g)$.
We compute the total expected cost of the state, including the estimated expected future cost via $\textsc{Estimator}(s'_g)$.
From this plan, we compute a set of possible \emph{alternate goal states} $G_\tau$ via a procedure $\textsc{SetOfAlternateGoalStates}$: each element $s_g \in G_\tau$ is a potential state in which the task $\tau$ is satisfied.
To save computation, the procedure $\textsc{SetOfAlternateGoalStates}$ is not exhaustive, and instead enumerates alternate placements only of objects manipulated during solution plan $\pi$, keeping only those that still solve the task $\tau$.
For example, in Fig.~\ref{fig:schematic}, for the first task, the immediate goal state is block \texttt{A} being in the red region and block \texttt{C} being in blue. One such alternate goal state could be to place block \texttt{A} in the red region and \texttt{C} in the white region.
We iterate over all $s_g \in G_{\tau}$ to find the alternative intermediate goal state that minimizes the total cost---computed via both the Fast Downward solver (Sec.~\ref{pddl}) and our \textsc{Estimator} for the anticipatory cost.

Similarly, for our \emph{preparation} tasks, for which a current task is not given, planning via Eq.~\eqref{eq:prep} involves minimizing the anticipatory cost over state space. 
This algorithm follows a hill-climbing optimization approach.
First, the algorithm is seeded with the initial state $s_0$. 
For 200 iterations, the robot randomly samples tasks from the task distribution and evaluates whether the anticipatory cost $V_{A.P.}$ improves along the way to solving it, computed using the trained model from Sec.~\ref{sec:learning}.
If the anticipatory cost of the resulting state is lower than that of the previously visited state, the robot chooses that state as the prepared state ($s^*_{A.P.}$) and repeats.

\section{Experiments and Results}

\begin{figure}
    \vspace{5pt}
    \hspace{5pt}
    \includegraphics[width=8cm]{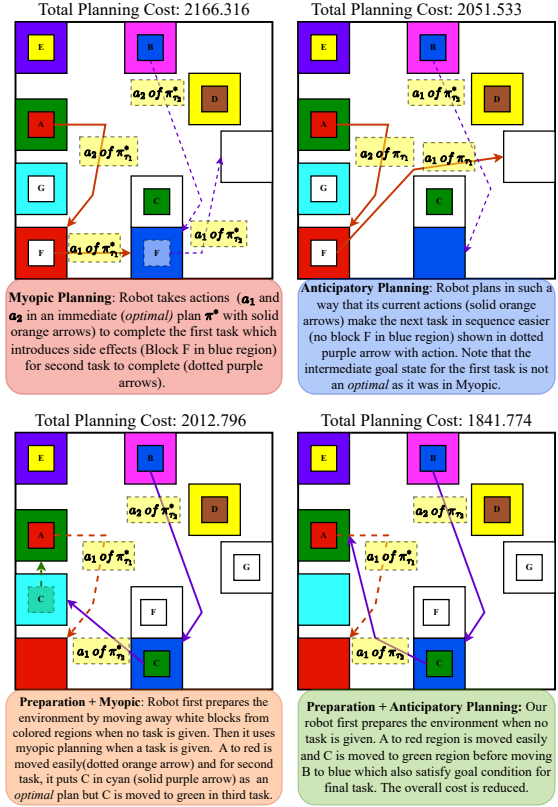}
    \caption{\textbf{Anticipatory Planning Reduces Total Planning Cost of a Sequence:} We provide the robot to solve the sequence of three tasks. We see the total planning cost is improved with anticipatory planning. The robot minimizes the overall cost according to Eq.~\eqref{eq:antplan} in which it estimates $V^*_{A.P.}$.}
    \centering
    \label{fig:bw}
    \vspace{-15pt}
\end{figure}

We evaluate our Anticipatory Planning approach on a 2D Blockworld environment, with procedurally generated worlds as shown in Fig.~\ref{fig:bw}.
We include experimental trials showing both (i) planning beginning from a random initial configuration of the environment and also (ii) the impact of \emph{preparation}: task-free anticipatory planning (as described in Sec.~\ref{sec:theory:prep}). 

For this environment, the task distribution is hand coded and has 20--25 pick-and-place-style tasks for each environment. Task specifications consist of placement directives for one or two blocks; for example, a task could involve moving only block A to the red region or moving block A and block B to the red and blue region respectively. The tasks are randomly sampled in a way that they are achievable in the environment and that only non-white blocks and non-white regions are involved. We also note that multiple blocks cannot be placed in one location. We use a uniform distribution over possible tasks. We then generate training data as described in Sec~\ref{sec:learning:data}.

For each environment, we evaluate over 100, 10-length task sequences, drawn according to the task distribution $\tau \sim P(\tau)$, each in 32 different randomly generated environments, for a total of 32,000 task executions.
For each trial, we evaluate performance using four planning approaches that leverage different aspects of our learning-augmented anticipatory planning approach:
\begin{LaTeXdescription}
  \item[Non-Learned Myopic Baseline (N.L. Myopic)] Classical planning via Fast Downward~\cite{helmert2006fast}, without any anticipatory planning cost. Plans only to minimize immediate cost $V^*_\tau(s_0)$.
  \item[Anticipatory Planning (A.P.)] Planning is augmented with estimates of the anticipatory planning cost. Planning seeks to reduce the overall cost of both the immediate task and a single future task, Eq.~\eqref{eq:antplan}.
  \item \textbf{Preparation + Myopic Baseline (Prep + N.L. Myopic)} Non-learned myopic planning, yet the robot is first allowed to \emph{prepare} the environment, Eq.~\eqref{eq:prep}.
  \item \textbf{Preparation + Anticipatory Planning  (Prep + A.P.)} Anticipatory planning, yet the robot is first allowed to \emph{prepare} the environment, Eq.~\eqref{eq:prep}.
\end{LaTeXdescription}


\begin{table}
\vspace{7pt}
    \begin{subtable}[h]{0.45\textwidth}
        \centering
        \begin{tabular}{cccc}
        \toprule
        Approach & Average Cost \\
        \midrule
        N.L. Myopic & 827.8 \\
        A.P. (ours) & \textbf{785.1}\\
        \bottomrule
      \end{tabular}
      \caption{Anticipatory Planning}
      \label{tab:ap}
    \end{subtable}
    \newline
    \vspace*{0.2 cm}
    \newline
    \begin{subtable}[h]{0.45\textwidth}
        \centering
        \begin{tabular}{cccc}
        \toprule
        Approach & Average Cost\\
        \midrule
        Prep (ours) + N.L.Myopic & 779.0 \\
        Prep (ours) + A.P. (ours) & \textbf{735.9} \\
        \bottomrule
        \end{tabular}
        \caption{Preparation: Task-Free Anticipatory Planning}
        \label{tab:prep}
     \end{subtable}
     \caption{Average cost per task in a 10-task sequence, averaged over 100 sequences in each of 32 Blockworld environments. }
     \label{tab:res}
     \vspace{-10pt}
\end{table}

We show our results in Table~\ref{tab:res}, which show that anticipatory planning \texttt{A.P.} (in the absence of preparation) reduces overall planning cost across all sequences by 5\%. \texttt{N.L. Myopic} has higher planning costs because the plans generated from the task will introduce side effects on subsequent tasks, increasing the cost of their plans.

We additionally evaluate scenarios in which we first allow both robots to \emph{prepare} their environment via task-free anticipatory planning, Eq.~\eqref{eq:prep}.
When allowed to prepare the environment in advance using estimates of the anticipatory cost $V^*_\text{A.P.}$, the performance of both planners improves.
After preparation, even the non-learned baseline \texttt{Prep+N.L. Myopic} shows improved performance by 5\%, since objects that could potentially clutter and obstruct its plans are moved out of the way in advance.
With preparation, our anticipatory planning approach \texttt{Prep+A.P.} improves overall performance by 11\% compared to the non-prepared, non-learned baseline \texttt{N.L. Myopic}.

\begin{figure}[t]
    \vspace{7pt}
    \includegraphics[width=8.5cm]{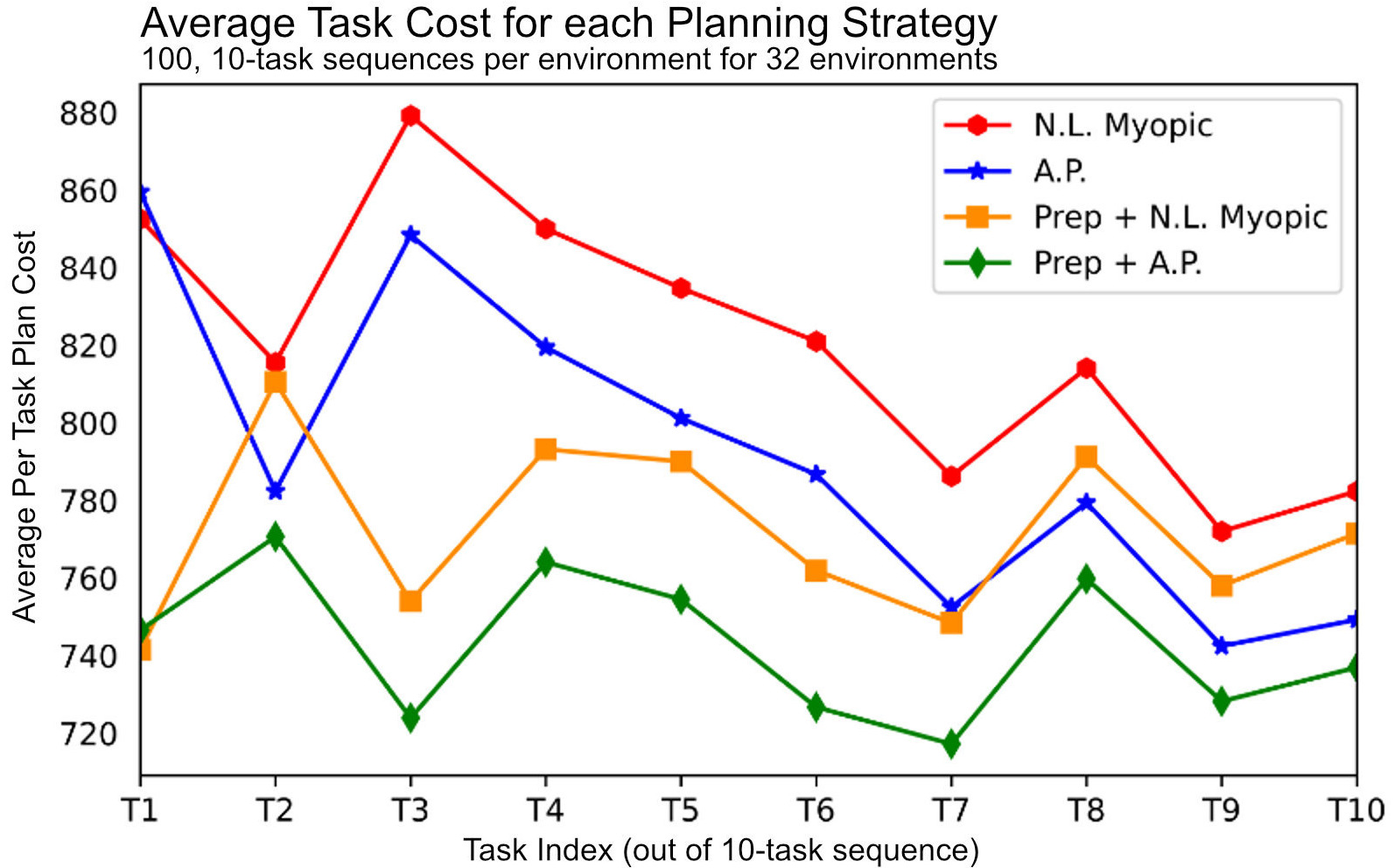}
    \caption{Average total cost of planning 100, 10-task sequences in each of 32 unique Blockworld Environments. Each task sequence is unique and randomly sampled from the set of task; the ``task index`` thus corresponds to how many tasks in each sequence have so far been completed.}
    \centering
    \label{fig:bw_curve}
    \vspace{-15pt}
\end{figure}

Critically, even in the absence of preparation, the task cost for our anticipatory planning approach (\texttt{A.P.}) gradually decreases as it accomplishes more tasks, suggesting that it is gradually \emph{preparing} or \emph{organizing} the environment, incrementally making it cheaper to plan. By the end of the sequences (at \texttt{T10} in Fig.~\ref{fig:bw_curve}), the average task cost approaches that of the initially prepared environment with anticipatory planning.

Furthermore, to qualitatively illustrate the benefits of both anticipatory task planning and preparation, we highlight one result scenario in Fig.~\ref{fig:bw}.
An example task sequence is given to the robot to complete one task at a time: \emph{[(\texttt{T1}) Move block \texttt{A} to the \texttt{red} region, (\texttt{T2}) Move block \texttt{B} to the \texttt{blue} region, (\texttt{T3}) Move block \texttt{C} to the \texttt{green} region]}.
\texttt{N.L. Myopic} first places block \texttt{F} in the \texttt{blue} region and block \texttt{A} in the \texttt{red} region to satisfy the goal condition. However, its placement of block \texttt{F} on the \texttt{blue} region introduces a side effect, so that the robot must subsequently take an additional step to complete the second task: moving block \texttt{F} to the \texttt{white} region.
Under our anticipatory planning approach, the white block \texttt{F} is moved directly to the \texttt{white} region, avoiding this expensive side effect at the expense of a small initial cost.



\section{Conclusion and Future Work}
In this work, we present Anticipatory Planning, an approach for learning-augmented planning that aims to reduce planning costs for long-lived household robots expected to accomplish sequences of tasks, yet given only one at a time.
Through the estimation of the \emph{anticipatory planning cost}, and thus the expected impact actions on potential future actions, our approach guides the robot towards behaviors that reduce overall planning cost.

Additionally, when a task is not given---a common scenario for household robots, which may not be constantly given active instruction---the robot is allowed to \emph{prepare} the environment (task free anticipatory planning) so as to reduce expected future costs when it is eventually given a task.


In future work, we seek to apply this approach to guide \emph{integrated} Task and Motion Planning, which would make our approach more broadly applicable in all manner of household domains in which the robot's implicit decisions can negatively impact future tasks and the state of its surroundings. \added[id=RD]{In such environments, one could use an existing dataset, such as the ALFRED task benchmark~\cite{shridhar2020alfred}, to provide tasks to define the task distribution.}

\section*{Acknowledgement}
We thank the Robust Robotics Group at MIT, Leslie Kaelbling, Nicholas Roy, Jana Kosecka, George Konidaris, Tom Silver, and Rohan Chitnis for their thoughtful feedback on this work. We acknowledge funding support from George Mason University.

\bibliographystyle{IEEEtran}
\bibliography{references.bib}
\end{document}